\documentclass[twocolumn]{article}

\pdfoutput=1

\usepackage[pdftex]{graphicx} \pdfcompresslevel=9

\usepackage{t1enc,dfadobe}

\usepackage{cite}

\usepackage[english]{babel}
\usepackage{abstract}
\usepackage{blindtext}
\usepackage{graphicx}
\usepackage{amsfonts}
\usepackage{amssymb}
\usepackage{caption}
\usepackage{subcaption}
\usepackage{url}
\usepackage{hyperref}

\title{Deep Learning Multidimensional Projections}
\date{}

\author{\parbox{\textwidth}{\centering Mateus Espadoto, Nina S. T. Hirata and Alexandru C. Telea}}

\makeatletter
\def\endthebibliography{%
  \def\@noitemerr{\@latex@warning{Empty `thebibliography' environment}}%
  \endlist
}
\makeatother

\begin{document}

\twocolumn[%
  \maketitle

  \begin{abstract}
  Dimensionality reduction methods, also known as projections, are frequently used for exploring multidimensional data in machine learning, data science, and information visualization. Among these, t-SNE and its variants have become very popular for their ability to visually separate distinct data clusters. However, such methods are computationally expensive for large datasets, suffer from stability problems, and cannot directly handle out-of-sample data. We propose a learning approach to construct such projections. We train a deep neural network based on a collection of samples from a given data universe, and their corresponding projections, and next use the network to infer projections of data from the same, or similar, universes. Our approach generates projections with similar characteristics as the learned ones, is computationally two to three orders of magnitude faster than SNE-class methods, has no complex-to-set user parameters, handles out-of-sample data in a stable manner, and can be used to learn any projection technique. We demonstrate our proposal on several real-world high dimensional datasets from machine learning.
  \end{abstract}
]


\section{Introduction}
\label{sec:intro}
Exploring high-dimensional data sets is a key task in many application domains such as statistics, data science, machine learning, and information visualization. The main difficulty encountered by this task is the large size of such data sets, seen both in the number of observations (also called samples) and number of measurements recorded per observation (also called dimensions, features, or variables). As such, high-dimensional data visualization has become an important separate field in information visualization (infovis)\,\cite{kehrer13_survey,maljovec15}.

Several techniques have been promoted for high-dimensional data visualization, including glyphs, parallel coordinate plots, table lenses, scatterplot matrices, dimensionality reduction methods, and multiple views linking the above visualization types. In this family of techniques, dimensionality reduction methods, also called projections, have a particular place: Compared to all other techniques, they scale much better in terms of both the number of samples and the number of dimensions they can accommodate (show) on a given screen space area. As such, projections have become the tool of choice for exploring data which has an especially high number of dimensions (tens up to hundreds) and/or in applications where the individual identity of dimensions is less important, as frequently met data science and machine learning applications. In the last decade, many projection techniques have been proposed\,\cite{maaten09_survey,nonato18_survey}. Among these, t-SNE\,\cite{maaten2008visualizing} is arguably one of the best known and most adopted in applications, given it creates projections with good visual segregation of similar-sample clusters. Yet, t-SNE comes with some downsides: It can be very slow to run on large data sets (thousands of observations or more), due to its quadratic nature; its parameters can be tricky to get right, which can lead to unpredictable results; and it lacks the capability of projecting out-of-sample data, which is useful when comparing several (time dependent) datasets\,\cite{rauber17_dyntsne,nonato18_survey}.

Work has been done to address the performance issue, such as tree-accelerated SNE\,\cite{vandermaaten14a}, H-SNE\cite{pezzotti2016}, A-SNE\cite{Pezzotti2017ApproximatedAU}, and UMAP\cite{mcinnes2018umap}. However, in general, there is no technique in the t-SNE class that jointly addresses scalability, stability, and out-of-sample handling. Moreover, t-SNE refinements are algorithmically not simple to understand and/or implement, which may limit their attractiveness. Such limitations are, to a large extent, shared by many other projetion techniques\,\cite{nonato18_survey}. Hence, a way to handle them jointly and independently on the projection technique of choice is of considerable interest.

To address these issues, we propose a learning-based approach to dimensionality reduction: We take any projection technique (considered suitable for an application at hand), run it on a small subset of the available data, train a deep neural network to learn the mapping from high to low dimensional space, and use the trained network to project the entire dataset, or similar datasets. Our method has the following contributions:

\noindent\textbf{Quality (C1):} We provide similar levels of visual segregation of data clusters as the learned projection, be it t-SNE or other;

\noindent \textbf{Scalability (C2):} We compute the projection in linear time in the number of dimensions and observations; practically, we project datasets of over hundred thousand observations and hundreds of dimensions in a few seconds using commodity hardware;

\noindent \textbf{Ease of use (C3):} Our method works without the need to set any complex parameters; our method is implemented using only widely accessible, open-source, infrastructure;

\noindent \textbf{Genericity (C4):} We can handle any kind of high-dimensional data, and can mimic the behavior of different types of existing projection techniques;

\noindent \textbf{Stability and out-of-sample support (C5):} Our method allows one to project additional data along a given (existing) dataset. Small changes in the data influence the resulting projection only slightly.

This paper is structured as follows. Section~\ref{sec:related} discusses related work on multidimensional projections. Section~\ref{sec:method} details our method. Section~\ref{sec:results} presents our results that support our claims outlined above. Section~\ref{sec:discussion} discusses our proposal. Section~\ref{sec:conclusion} concludes the paper.







\section{Related Work}
\label{sec:related}
We first introduce a few notations. Let $\mathbf{x} = (x^1,\ldots,x^n)$, $x^i \in \mathbb{R}, 1 \leq i \leq n$ be a $n$-dimensional ($n$D) real-valued observation or sample, and let $D=\{\mathbf{x}_i\}$, $1 \leq i \leq N$ be a dataset of $N$ samples. Let $\mathbf{x}^{j} = (x_1^j, \ldots, x_N^j)$, $1 \leq j \leq n$ be the $j^{th}$ feature vector of $D$. Thus, $D$ can be seen as a table with $N$ rows (samples) and $n$ columns (features or dimensions). A projection technique is a function
\begin{equation}
P : \mathbb{R}^N \times \mathbb{R}^n \rightarrow \mathbb{R}^N \times \mathbb{R}^q, 
\end{equation}
where $q \ll n$, and typically $q=2$. The projection $P(\mathbf{x})$ of a sample $\mathbf{x} \in D$ is a 2D point. Projecting a set $D$ yields thus a 2D scatterplot, which we denote as $P(D)$. 

\noindent\textbf{Dimensionality reduction:} Over the years tens of Dimensionality Reduction (DR) methods have been developed. These propose quite different trade-offs between the six desirable features listed in Sec.\ref{sec:intro}, as follows. For more extensive reviews of DR methods, and their quality features, we refer to\,\cite{hoffman02,maaten09_survey,engel12,sorzano14_survey,maljovec15,cunningham15_survey,xie17_survey}.

Probably the best known DR method is Principal Component Analysis\cite{jolliffe1986principal} (PCA), which has been used in several areas for many decades. It is a very simple algorithm with theoretical grounding in linear algebra. PCA is commonly used as preprocessing step for automatic DR on high-dimensional datasets prior to selecting a more specific DR method for visual exploration\,\cite{nonato18_survey}. PCA scores high on scalability (C2), ease of use (C3), predictability, and out-of-sample capability (C5). However, due to its linear and global nature, PCA lacks on quality (C1), especially for data of high intrinsic dimensionality, which is less than ideal for data visualization purposes.

Methods based on Manifold Learning, such as MDS\,\cite{torgerson58}, Isomap\,\cite{tenenbaum2000global} and LLE\,\cite{roweis2000nonlinear} and its variations\,\cite{donoho2003hessian,zhang2004principal, zhang2007mlle} try to reproduce in 2D the high-dimensional manifold on which data is embedded, and are designed to capture nonlinear structure in the data. These methods are commonly used in visualization since they generally yield better results than PCA (or similar global/linear methods) in terms of quality (medium-high C1). Unfortunately, those methods are harder to tune (low C3), do not have out-of-sample capability (C5), and generally scale poorly for large datasets (low C2).

A decade ago, the SNE (Stochastic Neighborhood Embedding) family of methods appeared, of which t-SNE\,\cite{maaten2008visualizing} is arguably the most popular member. A key praised feature of t-SNE is the ability to visually segregate similar samples in $D$ (C1), which can be useful as a preprocessing step on unsupervised learning set-ups, such as clustering. Despite its high scoring on the quality criterion (high C1), t-SNE can be very slow (low C2), since it has a complexity of $O(N^2)$ in sample count, is very sensitive to small changes in the data (low C5), can be very hard to tune (low C3) in order to get good visualizations, and does not have out-of-sample capability. There are attemps to improve t-SNE's performance, such as tree-accelerated t-SNE\,\cite{vandermaaten14a}, hierarchical SNE\,\cite{pezzotti2016}, and approximated t-SNE\,\cite{Pezzotti2017ApproximatedAU}. However, these methods require quite complex algorithms, and still largely suffer from the aforementioned sensitivity, tuning, and out-of-sample problems. More recently, UMAP (Uniform Manifold Approximation and Projection)\,\cite{mcinnes2018umap} appeared, advertised as a method which can generate projections with comparable quality to t-SNE (high C1) but much faster (high C2), and it has out-of-sample capability. Despite its advantages, it shares some disadvantages with t-SNE, namely, its susceptibility to small data changes (low C5) and parameter tuning difficulty (low C3).

\noindent\textbf{Deep learning:} Neural network approaches have been proposed for DR, such as Autoencoders\,\cite{hinton2006reducing,kingmaW13}, which aim to generate a compressed, low-dimensional representation on their bottleneck layers by training the network to reproduce its inputs on its outputs. Typically, autoencoders produce results comparable to PCA on the quality criterion (low C1). Also, it can be difficult to find a proper autoencoder architecture that yields good results for a specific dataset family (low C3). However, autoencoders are easily parallelizable (high C2), predictable, and provide out-of-sample capability (C5).

The ReNDA algorithm\,\cite{beckerLS17} is a very recent neural-based approach that uses two networks, improving on earlier work from the same authors. One network is used to implement a nonlinear generalization of Fisher's Linear Discriminant Analysis, using a method called GerDA; the other network is an Autoencoder used as a regularizer. According to the results of the original paper, the method scores well on predictability and has out-of-sample capability (C5). However, it requires labeled data, which none of the other algorithms in this study do. Also, the authors do not present results that show how scalable the method is (unknown C2).

Given the high success of t-SNE related to criterion C1, we will next focus mainly on methods in the SNE class that aim to satisfy C1 by construction. However, we will also consider other DR methods that are particularly successful for (some of) the named criteria.




\section{Method}
\label{sec:method}
Our proposal has a very simple structure: consider a data \emph{universe} ${\mathcal D}$, \emph{i.e.}, the union of all data sets created by a given application area, \emph{e.g.} all fashion images, all handwritten digit images, or all astronomical images related to a certain type of measurement. If we admit that there exists some specific structure of the data in such a universe, \emph{i.e.}, the data samples are not uniformly distributed along all dimensions, then a \emph{good} projection should capture well this data structure (which is often reflected in terms of segregating different data clusters in the visual space). We hypothesize that the way in which a given projection technique $P$ captures this data structure can be \emph{learned} by using a limited number of small training data sets $D \subset \mathcal{D}$ and their respective projections $P(D) \subset \mathbb{R}^2$.

Our proposal follows precisely this: Let $D_s$ be a randomly-selected subset of one or several datasets $D \subset{\mathcal D}$, and $P(D_s)$ be the corresponding projection of $D_s$. Let $P_{nn}$ be a neural network trained on $D_s$ aiming to mimic the behaviour of $P(D_s)$. Let $D_p = D \setminus D_s$ be the remaining data in $D$ to be projected by $P_{nn}$.

\begin{figure}[!htp!]
\centering
\includegraphics[width=0.48\textwidth]{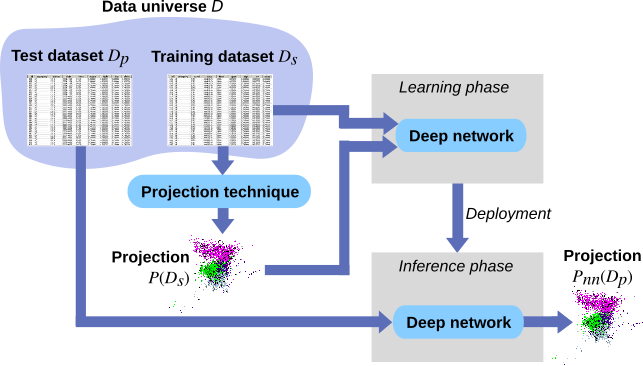}
\caption{Pipeline for learning projections (see Sec.~\ref{sec:method}).}
\label{fig:pipeline}
\end{figure}

Figure~\ref{fig:pipeline} presents our idea, which consists of three main steps -- creation of the training projection, training, and inference. The creation of the training projection is the process of projecting the subset $D_s$ by using a projection technique $P$. We use the projected subset $P(D_s)$ alongside the original high-dimensional $D_s$ to train a fully-connected neural network $P_{nn}$ to learn how to project high-dimensional data. Once $P_{nn}$ is trained, we use it to project the remaining points $D_p$ of $D$.
Moreover, and by extension, we use $P_{nn}$ to also project different datasets from the same universe ${\mathcal D}$ as $D$.

After some empirical testing, varying the number of layers and the number of elements in each layer, we defined the architecture for $P_{nn}$ as having three fully-connected hidden layers, with 256, 512, and 256 units respectively, using ReLU activation functions, followed by a 2-element layer which uses the sigmoid activation function to encode the 2D projection, scaled to the interval $[0, 1]^2$ for implementation simplicity. We initialize weights with the He uniform variance scaling initializer\,\cite{he2015delving}, and bias elements using a constant value of 0.0001, which showed good results during testing. We use the Adam\,\cite{kingma2014adam} optimizer to \emph{train} $P_{nn}$ for up to 200 epochs on an ``early stopping'' setup, which stops the training automatically on convergence, defined as the moment when the validation loss stops decreasing. In practice, no more than 60 epochs were needed to achieve convergence, the average being 30 epochs (for full details, see Sec.~\ref{sec:q_how_much}). The cost function used was mean squared error, which showed better convergence speed during testing than other common cost functions, such as mean absolute error and log hyperbolic cosine (logcosh).

We \emph{test} $P_{nn}$ by comparing the projections it delivers on $D_p$ (unseen data during training) with the ground truth $P(D_p)$ obtained by running the projection $P$ we desire to imitate on $D_p$. For this, we use labeled data: We compare both the visual segregation of same-label samples as given by $P_{nn}$ and $P$ (qualitative comparison) and the neighborhood hits\,\cite{paulovich2008least} of $P_{nn}(D_p)$ and $P(D_p)$, which is a well-known metric for assessing the quality of projections (quantitative comparison). Note, however, that labels are not used anywhere during training or computing the ground truth projection.


\section{Results}
\label{sec:results}
We next show how our proposal covers the initial requirements listed in Sec.~\ref{sec:intro}. For this, we structure our evaluation into several tasks. We compare our results with those produced by several well-known projection techniques (t-SNE, UMAP, PCA, Isomap, and LLE). We use a range of publicly available real-world datasets that have many observations and dimensions, exhibit a non-trivial data structure, and come from different application domains, as follows:

\noindent\textbf{MNIST\,\cite{lecun2010mnist}:} 70K observations of handwritten digits from 0 to 9, rendered as 28x28-pixel grayscale images, flattened to 784-element vectors;

\noindent\textbf{Fashion MNIST\,\cite{xiao2017online}:} 70K observations of 10 types of pieces of clothing, rendered as 28x28-pixel grayscale images, flattened to 784-element vectors;

\noindent\textbf{Cats and Dogs\,\cite{elson2007asirra}:} 25K images of varying sizes divided into two classes (cats, dogs). We used the Inception V3\,\cite{szegedy2016rethinking} Convolutional Neural Network (CNN) pre-trained on the ImageNet data set\,\cite{imagenet} to extract features of those images, yielding 2048-element vectors for each image;

\noindent\textbf{IMDB Movie Review\,\cite{imdb}:} 25K movie reviews from which 500 features were extracted using TF-IDF\,\cite{salton86}, a standard method in text processing;

\noindent\textbf{CIFAR-10 and CIFAR-100\,\cite{krizhevsky2009learning}:} 60K 32x32-pixel color images in 10 and 100 classes, respectively. We used the DenseNet\,\cite{densenet} CNN pre-trained on the ImageNet dataset to extract features of those images, yielding 1920-element vectors for each image.

For each dataset, the split between training and test sets varies for each experiment and is explained in detail next in each task-specific section.

\subsection{Training effort}
\label{sec:q_how_much}
\begin{figure}[!htp!]
    \centering
      \includegraphics[width=0.49\textwidth]{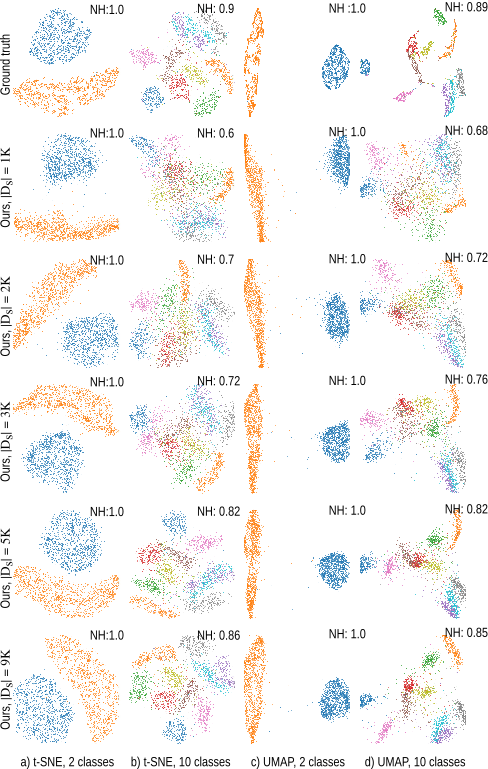}
\caption{Top row: MNIST dataset, 10K sample projections of 2 and 10 classes created by t-SNE and UMAP. Next rows: Projections done by our method using varying training set sizes $|D_s|$.}
    \label{fig:train_mnist}
\end{figure}

\begin{figure}
    \centering
    \includegraphics[width=0.48\textwidth]{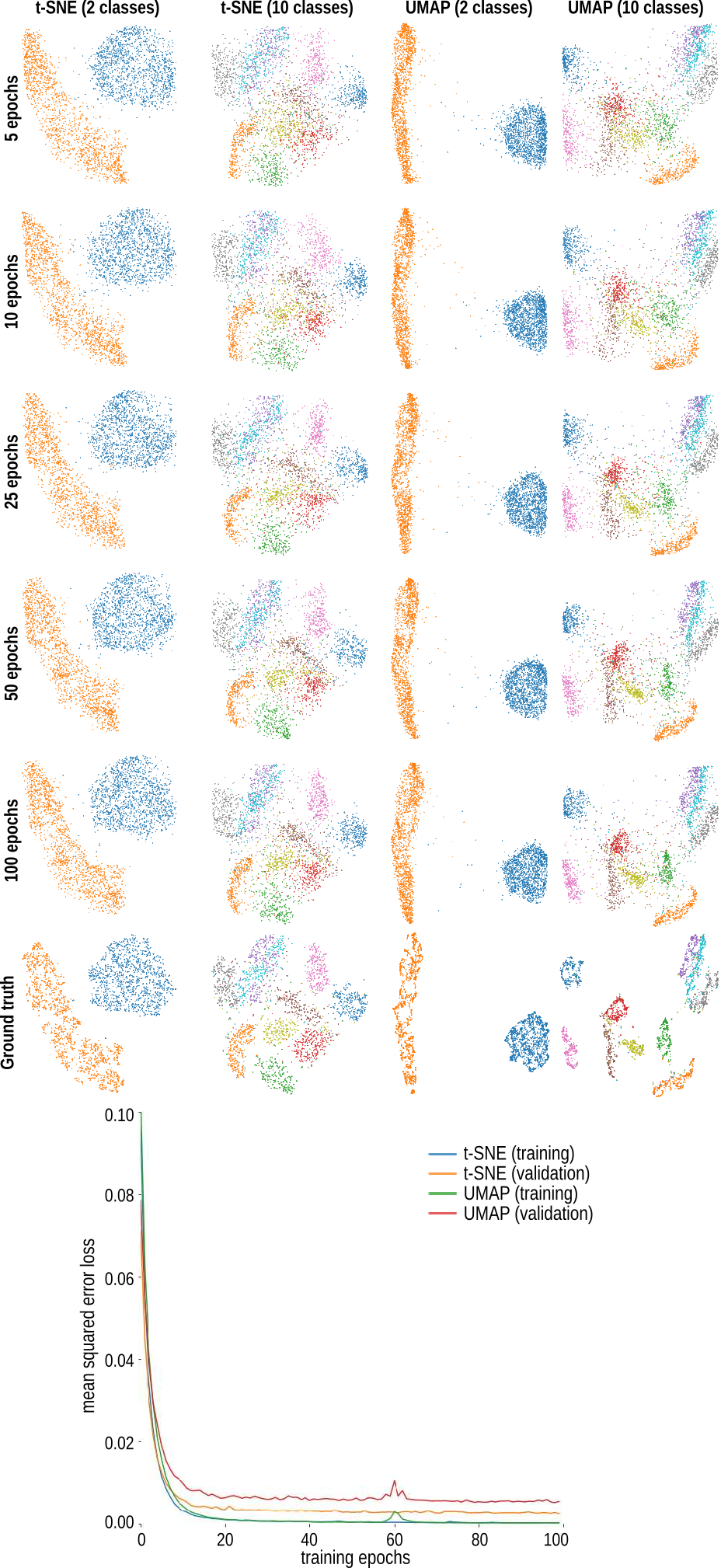}
\caption{Ground truth: MNIST dataset, 3K sample projections of 2 and 10 classes respectively created by t-SNE and UMAP. Rows above: Projections done by our method using varying numbers of epochs. Bottom graph: loss as function of number of training epochs, during both training and validation.}
    \label{fig:epochs_fmnist}
\end{figure}

It is important to assess what our method needs (training-data-wise and training-effort-wise) to reach the quality of the training projection, or close to that. Figure~\ref{fig:train_mnist} shows t-SNE and UMAP projections of subsets of the MNIST dataset with two and ten classes, respectively, alongside our method's results. We used   training sets $D_s$ of varying sizes, all randomly and indepedently sampled from the MNIST dataset. We included the two-class selection (digits 0 and 1) since we know that images for these digits are quite different. Hence, the obtained projections should clearly separate samples from these two classes. For the two-class case, we see that our method yields practically the same results as the ground truth methods (t-SNE and UMAP, respectively), already when using 1K training samples. For the 10-class case, we obtain very similar results starting from roughly 5K training samples. Separately, Figure~\ref{fig:epochs_fmnist}(top) shows, for both t-SNE and UMAP, how the quality improves for a fixed training set (3K samples) as we increase the number of training epochs. The bottom graph in the same figure shows how the loss (cost) decreases \emph{during training} as we increase the number of epochs (blue and green curves for t-SNE and UMAP, respectively). The orange and red curves (for t-SNE and UMAP, respectively) shows what the loss is for the \emph{validation} set for a network trained for a given number of epochs. As visible, all curves converge quite quickly -- of course, the validation loss is a bit larger than the training loss. We can thus use these curves in practice to find how many training epochs we need for a desired maximal loss. Conversely, we can fix a preset maximal loss (in practice, 0.005) and compute the number of training iterations required for it. Table~\ref{tab:epochs} shows the resulting numbers of training epochs required. This justifies the maximal preset of 60 training epochs (and its average of 30 epochs) mentioned in Sec.~\ref{sec:method}.

\begin{table}[htb]
    \centering
    \small
    \caption{Relationship between number of training samples and number of epochs needed to obtain convergence,  MNIST data set.}
    \label{tab:epochs}
\begin{tabular}{l | c | c | c }
    Projection & Classes & Samples & Epochs \\ \hline
    t-SNE      & 2  & 1000    & 57     \\
    t-SNE      & 2  & 2000    & 30     \\
    t-SNE      & 2  & 3000    & 50     \\
    t-SNE      & 2  & 5000    & 32     \\
    t-SNE      & 2  & 9000    & 24     \\ \hline
    t-SNE      & 10 & 1000    & 49     \\ 
    t-SNE      & 10 & 2000    & 33     \\
    t-SNE      & 10 & 3000    & 31     \\
    t-SNE      & 10 & 5000    & 21     \\
    t-SNE      & 10 & 9000    & 13     \\ \hline
    UMAP       & 2  & 1000    & 44     \\
    UMAP       & 2  & 2000    & 21     \\
    UMAP       & 2  & 3000    & 31     \\
    UMAP       & 2  & 5000    & 28     \\
    UMAP       & 2  & 9000    & 42     \\ \hline
    UMAP       & 10 & 1000    & 31     \\
    UMAP       & 10 & 2000    & 30     \\
    UMAP       & 10 & 3000    & 33     \\
    UMAP       & 10 & 5000    & 23     \\
    UMAP       & 10 & 9000    & 21     \\ \hline
\end{tabular}
\end{table}

\subsection{Capturing the structure of different datasets}
\label{sec:q_how_good}
\begin{figure*}[!htp!]
    \centering
      \includegraphics[width=1.0\textwidth]{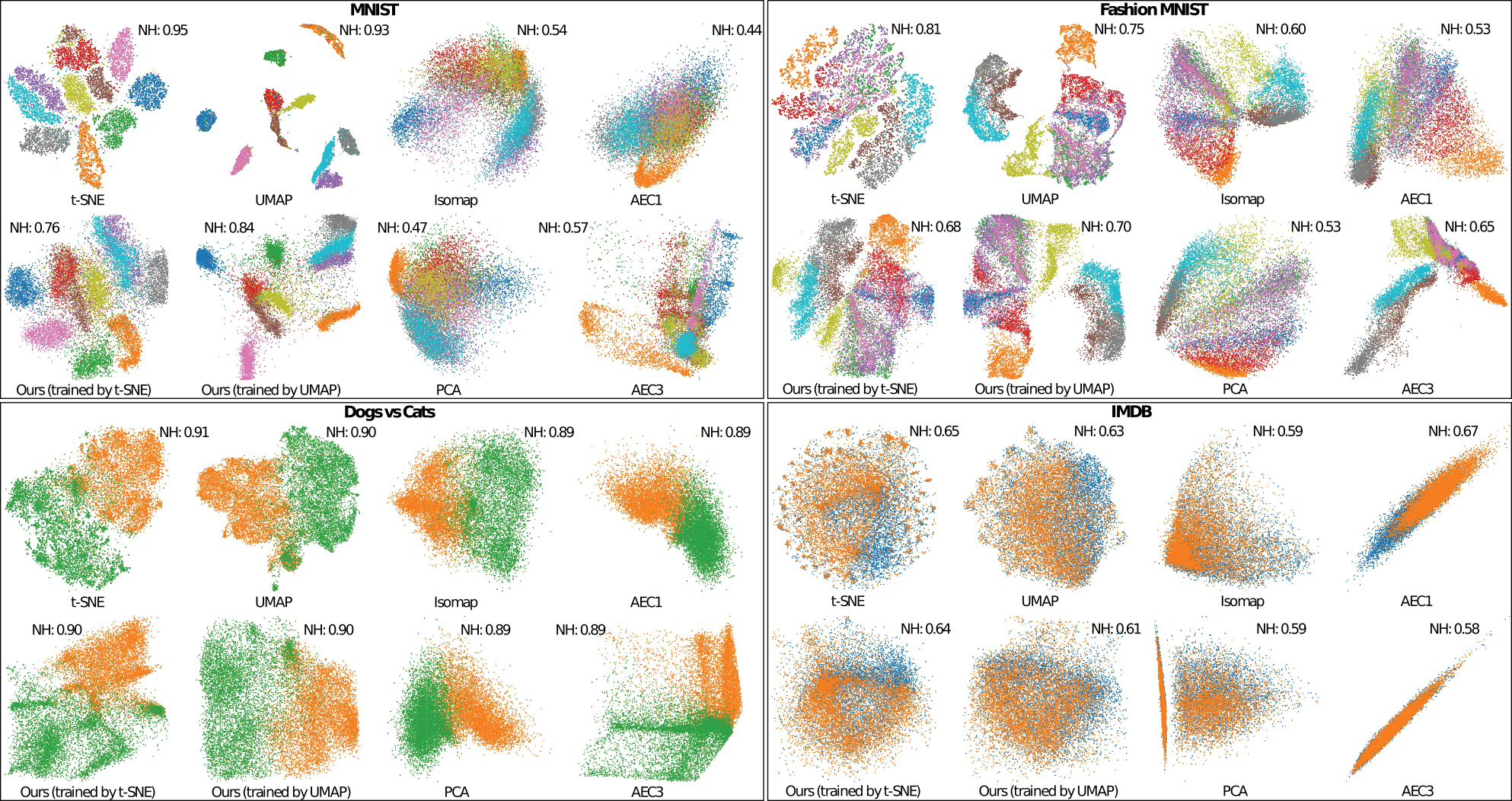}
    \caption{Projections (15K samples) created by several existing techniques (t-SNE, UMAP, Isomap, single-layer Autoencoder (AEC1), PCA, and 3-layer Autoencoder (AEC3)) on four different datasets. For comparison, our method, trained with t-SNE and UMAP, respectively.}
    \label{fig:train_all}
\end{figure*}

In dimensionality reduction, the ability to segregate clusters highly depends on the kind of input data and kind of projection technique used. To assess this ability for our technique, we compare it with t-SNE, UMAP, Isomap, PCA, and two variants of Autoencoders (having respectivey 1 and 3, layers), on 15K observation subsets of MNIST, Fashion MNIST, Cats and Dogs, and IMDB (Fig.~\ref{fig:train_all}). We trained our method to mimic t-SNE and UMAP, respectively, using 5K training samples. We first see that our method captures very well the visual data structure shown by t-SNE and UMAP. Separately, we see that PCA, Isomap, and Autoencoders, used in the canonical way, \emph{i.e.}, to project the data without any training, cannot capture well this structure. Hence, learning how to mimic a good projection that captures well the data structure (\emph{i.e.}, t-SNE or UMAP) is important for such datasets, and our method succeeds in doing so, clearly better than `vanilla' PCA, Isomap, and Autoencoders.

\subsection{Stability and out-of-sample data}
\label{sec:q_how_predictable}
We define \emph{stability} of a projection as the relation between the \emph{visual} changes in $P(D)$ related to \emph{data} changes in $D$. Ideally, a stable projection technique should not change $P(D)$ if $D$ does not change at all, regardless of changes in parameters of the algorithm $P$; and conversely, when $D$ changes, \emph{e.g.} as new samples are added, then the old samples should stay in $P(D)$ as close as possible to their original locations. This way, the user can relate changes in $P(D)$ to actual data changes. For a similar reasoning applied to different infovis algorithms, \emph{i.e.}, treemaps, see\,\cite{vernier18}. Hence, stability and out-of-sample capabilities are closely related. Doing this is not trivial. Many projection techniques use a random initialization, which means they can create quite different results for the same dataset $D$. Moreover, small parameter changes, \emph{e.g.}, perplexity for t-SNE, or choice of control points for LAMP, to mention just a few, can yield large changes in $P(D)$\,\cite{wattenberg_tsne}. Dynamic t-SNE corrects such effects up to a certain level, but comes with additional complexity and significant computational costs\,\cite{rauber17_dyntsne}.

\begin{figure*}[!htp!] 
    \centering
    \includegraphics[width=0.8\textwidth]{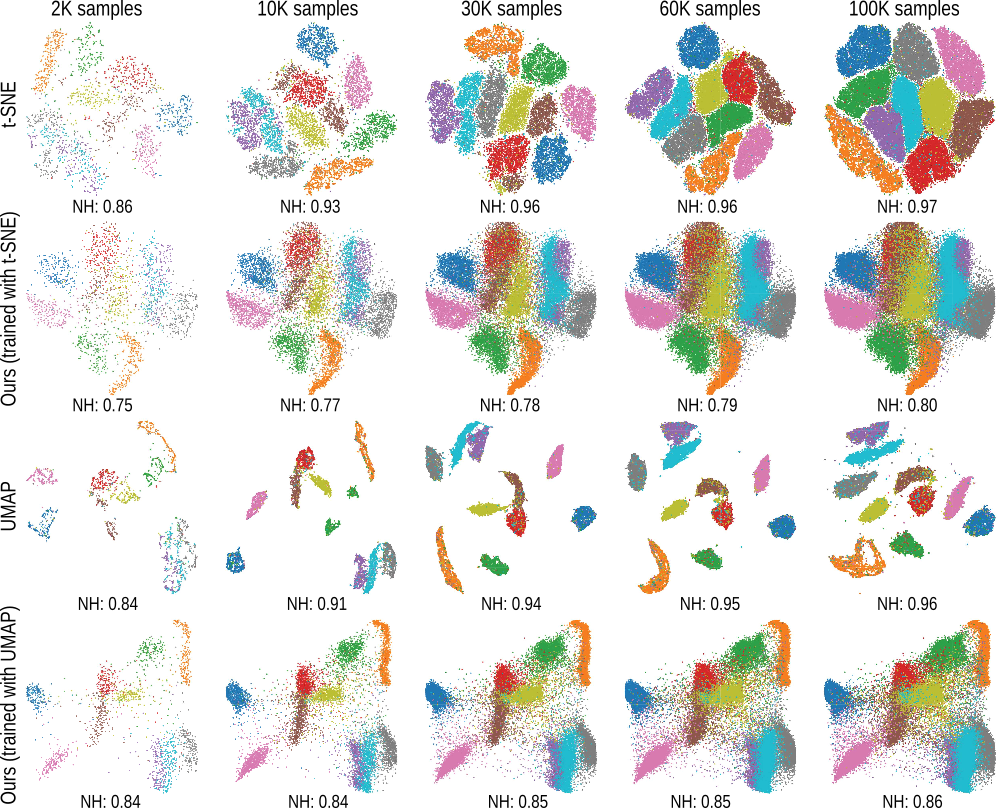}
    \caption{Projections of various number of samples from the MNIST dataset using different projection methods.}
    \label{fig:stab_mnist}
\end{figure*}

Figure~\ref{fig:stab_mnist} shows projections of the MNIST data set with increasing number of samples created by t-SNE, UMAP, and our method trained with 5K observations from t-SNE and UMAP respectively. We see that our method is very stable, albeit noisier, with clusters being rendered in the same places and only getting denser as more samples are added. Both t-SNE and UMAP show clusters in different placess as samples are added, which is not desirable for maintaining the user's mental map of the data. Hence, we argue that our method proposes a good trade-off between stability (and out-of-sample capability) \emph{vs} noisiness.

\subsection{Learning different projections}
\label{sec:q_how_generic_others}
\begin{figure*}[!htp!]
    \centering
        \includegraphics[width=1.0\textwidth]{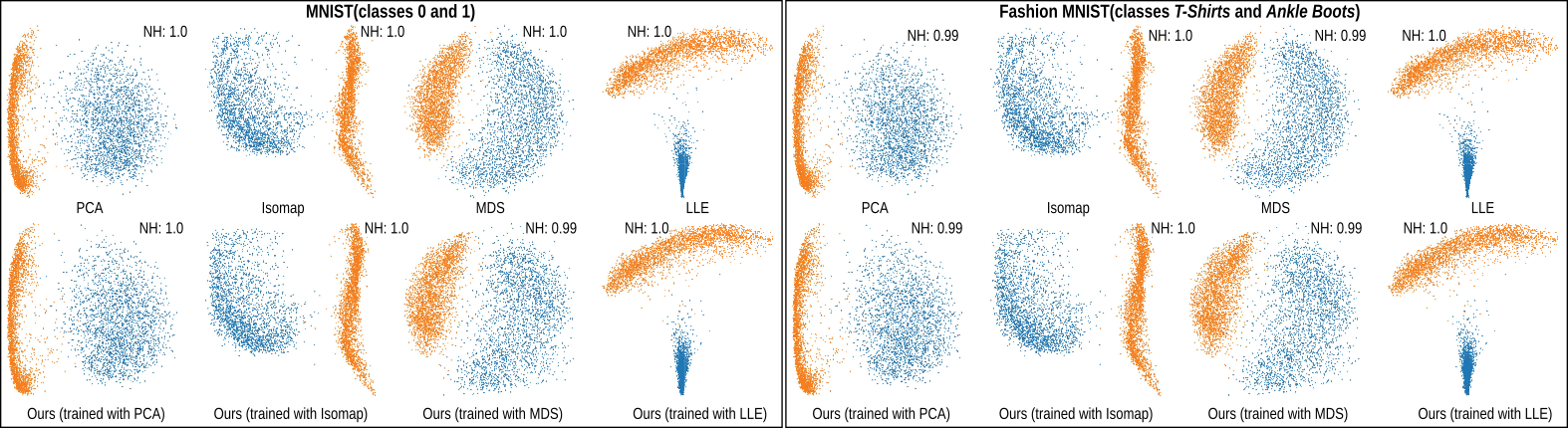}
    \caption{Learning different projections. Top row: Projections of 5K samples made with PCA, Isomap, MDS, and LLE. Bottom row: projections of different 5K samples, same datasets, created by our method trained on the data and projections from the top row.}
    \label{fig:others}
\end{figure*}

Our method can learn equally easily other projections than t-SNE or UMAP. Figure~\ref{fig:others} shows this by training our method to mimic PCA, Isomap, MDS, and LLE, for the MNIST and Fashion MNIST datasets. We used 5K training samples, and projected a different set of 5K samples from each of these two datasets. We see that our method can closely reproduce the patterns created by these four quite different projection techniques. Note that this does not imply that these projections are optimal in any sense of the word. Rather, our goal here is to show that our method can capture the style of different projection techniques. Also, note that there is nothing specific in the implementations of PCA, Isomap, MDS, or LLE in our approach: We claim, although we cannot (of course) formally prove, that we can learn any projection in the same way, since our method does not use any specifics of the learned projection technique.

\subsection{Computational scalability}
\label{sec:q_how_scalable}
\begin{figure}[!htbp]
\centering
\includegraphics[width=0.5\textwidth]{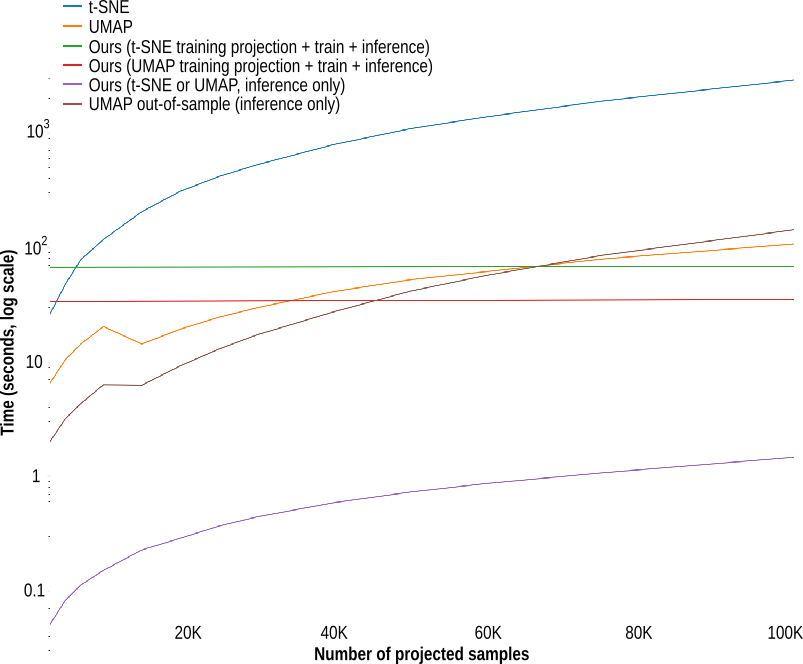}
\caption{Time to project varying number of samples, MNIST data set, oversampled to 100K observations (log time scale).}
\label{fig:time_mnist}
\end{figure}

One of our main goals was to create a projection technique which scales well to large datasets (C2, Sec.~\ref{sec:intro}). To analyze this, Fig.~\ref{fig:time_mnist} shows a time comparison between t-SNE, UMAP, and our method when projecting datasets of various sizes. We trained our method with 5K samples, in line with training-set sizes found to be sufficient in earlier experiments (Sec.~\ref{sec:q_how_much}). We first compare the performance of our \emph{end-to-end} method, \emph{i.e.}, considering the creation of the training projection, the training itself, and the inference --  see green and red lines in Figure~\ref{fig:time_mnist} for t-SNE and UMAP respectively. As visible, our end-to-end method already runs much faster than t-SNE starting from about 5K samples, and much faster than UMAP from about 30K samples, respectively. However, note that this is a worst-case scenario: In practice, one would train once on a given data universe $\mathcal{D}$ and project many times on the same $\mathcal{D}$. Hence, we next show the inference (projection) times only (purple curve). These times are identical for the learned t-SNE and UMAP, since we use the same network architecture for both. We see that our method is two to three orders of magnitude faster than t-SNE and about two orders of magnitude faster than UMAP. Finally, we consider UMAP's out of sample capability (see Sec.~\ref{sec:related} for details): We run UMAP on our training set, which makes it learn a function to transform the high-dimensional data to 2D. Note that this is completely different from our deep learning -- it is a particular feature of UMAP's implementation, not shared by other projection techniques. Next, we let UMAP use this learned function to project the test set. The inference only (projection) time of UMAP is shown by the brown curve. As expected, UMAP is faster in this case than when it needs to execute the entire project from scratch (orange curve), with a break-even point around 65K samples. However, even in this case, our method is one to two orders of magnitude faster than UMAP.

\subsection{Projecting unrelated data}
\label{sec:q_how_generic_tl}
So far, we showed that our method can learn from a subset of a given dataset $D$ to project unseen samples from the same $D$. When one avails of a large dataset $D$, which consistently samples a given problem domain, we can thus train once and next use the trained network many times to cover all applications within that domain. However, we can have the case when one does not readily have such a comprehensive $D$ for training, or does not want to spend the time to create expensive training projections for large $D$ datasets. The question is then: How can we \emph{reuse} the training done on a given type of data from some universe $\mathcal{D}$ (for which we have, for instance, sufficient training samples) to generate a network able to project data from a related, but still different, universe $\mathcal{D}'$, using a small number of samples from $\mathcal{D}'$?

\begin{figure*}[hp]
    \centering
        \includegraphics[width=0.86\textwidth]{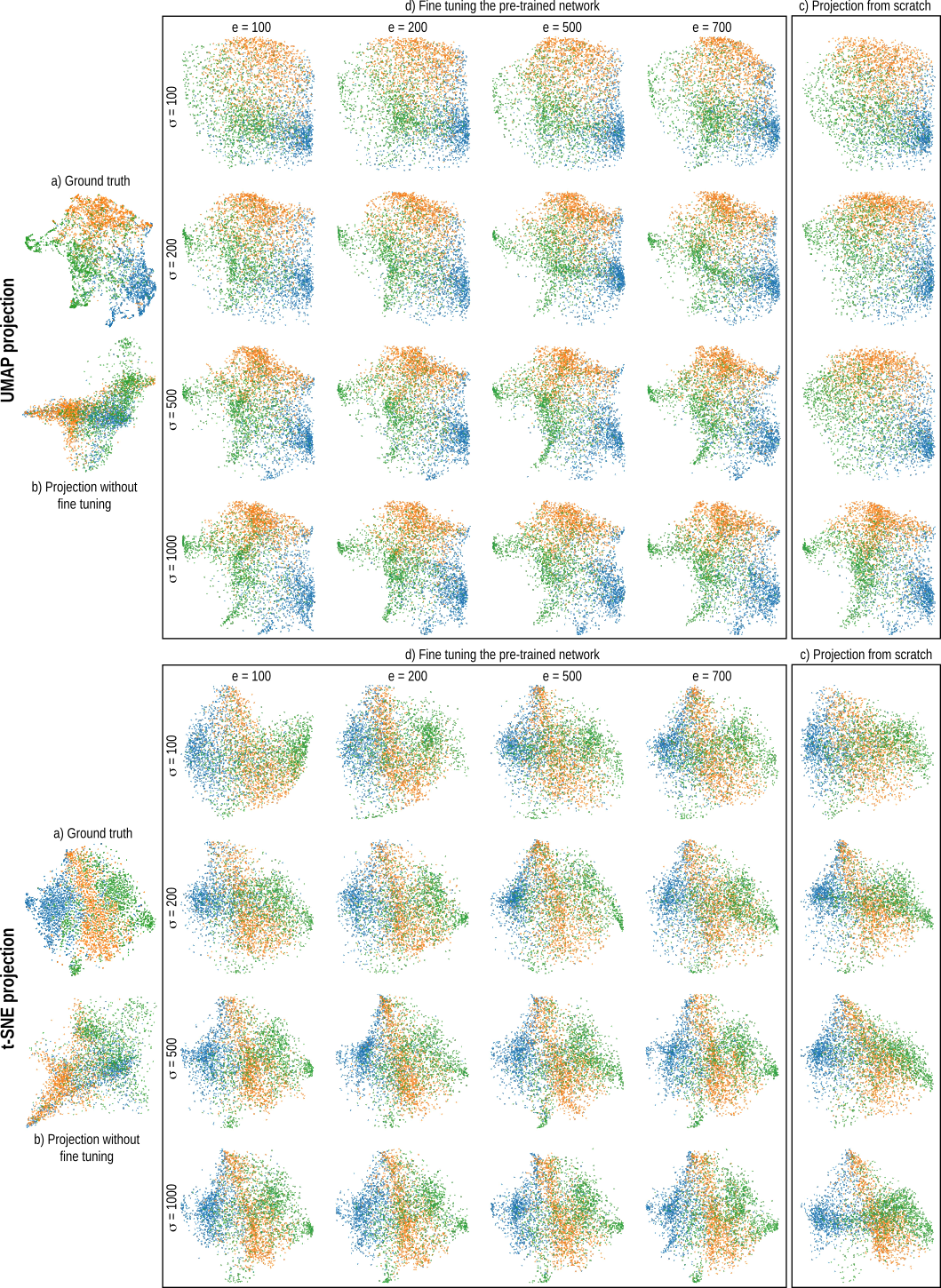}
\caption{Projecting data (4K samples) by fine tuning pre-trained networks mimicking UMAP and t-SNE. a) Test projection. b) Inference by pre-trained network without any fine tuning. Training uses 2K samples from universe $\mathcal{D}$. c) Projections made by our method trained from scratch from $\sigma$ samples from the new universe $\mathcal{D}'$. d) Projections made by with fine-tuning the pre-trained network with varying numbers of training epochs $e$ and using different numbers $\sigma$ of samples from $\mathcal{D}'$.}
    \label{fig:transfer_cifar}
\end{figure*}

To test the limits of training extrapolation, we conducted the following experiment. We trained our method using UMAP and t-SNE projections of 2K observations from CIFAR-10 (classes \emph{Airplane}, \emph{Frog} and \emph{Truck}), which constitutes thus a sampling of $\mathcal{D}$, the universe of natural images of vehicle and animal shapes. Next, we used the trained network to project 4K observations from CIFAR-100 (classes \emph{Trees}, \emph{Large Carnivores} and \emph{Vehicles 2}), which constitutes a sampling of $\mathcal{D}'$ -- a universe related, but not identical to, $\mathcal{D}$. We selected these classes because they contain images that are similar perceptually between the two universes $\mathcal{D}$ and $\mathcal{D}'$, with the goal of checking the capability of generalization of our method. Note however that $\mathcal{D}$ and $\mathcal{D}'$ are quite different: While both contain images, these are of quite different kinds, and acquired by different procedures. Figure~\ref{fig:transfer_cifar} shows the obtained results. First, we show the projections obtained by directly reusing the network trained on $\mathcal{D}$. As visible, the results are quite far from the ground truth (classical t-SNE and UMAP projections). This confirms that $\mathcal{D}$ and $\mathcal{D}'$ are, indeed, quite different, so extrapolating the training is not easy. We next consider training the network from scratch, using a small number of 100 to 1000 samples from $\mathcal{D}'$, mimicking the situation when the user has only few available data from $\mathcal{D}'$ to train on. To investigate the effect of fine tuning, we use 100 to 700 training epochs for each set of added samples. In brief, the procedure is very similar to \emph{transfer learning}\,\cite{pan10_survey}. As visible, training from scratch works reasonably well only when a large number of samples from $\mathcal{D}'$ are available. However, when using many training epochs, the fine-tuning yields similar results to the ground truth even for as few as 100 samples from $\mathcal{D}'$. This indicates that it is possible to create pre-trained \emph{domain-specific projection networks} for some data universe and then simply fine tune these for a different universe with a small number of samples from the latter.

\section{Discussion}
\label{sec:discussion}
We next discuss how our proposal meets the requirements introduced in Sec.~\ref{sec:intro}, but also the observed limitations.

\noindent \textbf{Quality (C1):} We showed that our method achieves very similar visual cluster separation patterns to projections well-known to perform well in this area, namely t-SNE and UMAP, on six challenging multidimensional datasets having hundreds of dimensions, often used as benchmarks in machine learning. In quantitative terms, our projections yield a lower quality as seen in terms of neighborhood hit (NH) values, roughly 12\% lower (see NH values for examples in the figures in the paper). However, such lower values do not affect the ability of our projection to show similar levels of cluster separation to t-SNE and/or UMAP, which is often one of the key reasons why they are used in practice. Also, our method yields both better visual cluster separation and higher NH values as compared to other known projection techniques such as autoencoders, Isomap, and PCA on difficult datasets (\emph{e.g.} Fig.~\ref{fig:train_all}).

\noindent \textbf{Scalability (C2):} Even when considering training, our method is roughly one order of magnitude faster than t-SNE and roughly 5 times faster than UMAP for more than roughly 30K samples. However, as explained in Sec.~\ref{sec:q_how_scalable}, this is a worst case, since one typically trains once and infers many times on a given data universe. For such cases, our method is more than two, respectively three \emph{orders of magnitude} faster than t-SNE, respectively UMAP, and allows projecting data of hundreds of thousands of samples in subsecond time. The complexity of our method is linear in the number of observations and dimensions. Besides t-SNE and UMAP, our method is actually also faster than other projection methods such as Isomap and LAMP.

\noindent \textbf{Ease of use (C3):} During inference, our method simply executes a trained neural network, which requires no parameter setting. There is no need for guessing the `right' values of parameters such as t-SNE's perplexity\,\cite{wattenberg_tsne}. During training, the only free parameter to be set is the maximal loss or, alternatively, number of training epochs. The two are related, see Sec.~\ref{sec:q_how_much}. As also explained there, a preset of 60 training epochs yielded a loss of 0.005, \emph{i.e.}, practical convergence, for all examples we considered.

\noindent \textbf{Genericity (C4):} Our method can learn the behavior of any type of projection technique. We provided examples in Sec.~\ref{sec:results} showing this for t-SNE, UMAP, MDS, Isomap, LLE, and PCA. All that is needed to learn is a number of samples from the data universe of interest, and their 2D coordinates computed by the desired projection technique. No other aspects or parameters of the training or inference process are projection-technique specific -- that is, projections to be learned can be seen as black boxes. Moreover, no restrictions exist in terms of the dimensionality $n$ of the input data and the dimensionality $q$ of the projected data. While we demonstrated our approach only for $q=2$ (2D projections), which are the most commonly used in infovis, producing higher-dimensional, \emph{e.g.}, 3D projections\,\cite{coimbra16}, is equally easy. Such projections are preferred in certain cases as they can preserve the original data structure better than 2D projections\,\cite{sanftmann12,coimbra16}. So far, we have only considered projecting quantitative data. However, extending our approach to handle categorical data is straightforward by using \emph{e.g.} one-hot encoding or similar techniques\,\cite{potdar17}.

\noindent \textbf{Stability and out-of-sample support (C5):} These two issues are strongly interconnected, and actually also connected with the question of how far our networks can generalize what they learn. Let us detail. As outlined in Sec.~\ref{sec:method}, we take a training set $D_s$ which is supposed to represent well the overall data distribution in a given so-called data universe $\mathcal{D}$, \emph{i.e.}, datasets related to a particular application, such as all handwritten digits, all human face images, all patients in a given population, all street views, and similar. Our approach learns how to project data in $\mathcal{D}$ based on training projections of data in $D_s$. Hence, the better $D_s$ represents the variability of data in $\mathcal{D}$, the better will our projections mimic actual projections of the same data. Given that neural netwok inference works deterministically, out-of-sample support is stable in the sense that the same data items (in a dataset $D \subset \mathcal{D}$) are projected to the same locations, which is not the case for many projection methods such as t-SNE, UMAP, and LAMP, to mention just a few. Separately, given the dense structure of the fully-conneted network we use (which averages activations from multiple units in an earlier layer to determine those of the current layer), our approach is stable in the sense that small changes in an input dataset yield only small changes in the resulting projection (see example in Sec.~\ref{sec:q_how_predictable}). Again, this result is far from evident for many existing projection techniques.

However, our results also show that there is a trade-off between this inherent stability and out-of-sample support and the quality (in terms of cluster separation) of the resulting projections. Globally put, our projections show fuzzier, or less sharply separated clusters, than corresponding t-SNE projections. This is the trade-off needed to provide stability: Our method cannot project samples as `freely' as \emph{e.g.} t-SNE, since it needs to behave deterministically; on the other hand, this ensures that the same location in $n$D space projects to the same place in $q$D space, which is not the case for t-SNE. A similar trade-off between stability and quality exists actually also for dynamic (t-SNE) projections\,\cite{rauber17_dyntsne}.

Related to the last point above, the question arises of \emph{how far} can our approach generalize, or, how densely do we need to sample an universe $\mathcal{D}$ by the training set $D_s$ to create good projections. This is an open question in machine (and deep) learning in general. Yet, we can make the following practical points. First, for the types of (non-trivial) data universes we consider in our evaluations, a few thousands of samples yield already high-accuracy results. Secondly, the larger $\mathcal{D}$ is, the larger (and better spread) the training set $D_s$ needs to be. Section~\ref{sec:q_how_generic_tl} outlines the limits of this extrapolation: The farthest away is $\mathcal{D}$ spread from the training set $D_s$, the more will our projections differ from the actual ground-truth projections obtained using classical projection methods. Again, this is not a surprise, but a well-known fact in machine learning. We argue that this is not a problem \emph{in practice} when using projections. Indeed: In all cases we are aware of, researchers typically work for a reasonable amount of time on a \emph{given}, and fixed, data universe $\mathcal{D}$. Hence, they can once train a network from a comprehensive $D_s \subset \mathcal{D}$, after which they can use the network with no changes for data in the same $\mathcal{D}$. Moreover, for cases where one targets a new data unverse $\mathcal{D}'$, for which obtaining a comprehensive training set $D_s$ is expensive, the transfer-learning-like approach in Sec.~\ref{sec:q_how_generic_tl} can be used. As shown in Sec.~\ref{sec:q_how_generic_tl}, one can fine-tune a pre-trained network (on widely available data from a related universe $\mathcal{D}$) with as few as hundreds of samples from $\mathcal{D}'$.

\section{Conclusion}
\label{sec:conclusion}
We have presented a new method for creating projections of high-dimensional data using a machine learning approach. Based on a small number of projections of a subset of samples from a given data universe, obtained using any user-chosen projection technique, we train a neural network to mimic the (2D) projection output, and next use the network to infer projections of unseen data from the same universe. Our method can mimic the quality and visual style of a wide range of established projection techniques, including the well-known visual cluster separation provided by SNE-class methods; is orders of magnitude faster than such methods; has a single parameter (with documented preset) for training, and no parameters for inference; can handle datasets of any (quantitative) kind and dimensionality; and delivers inherent stability and out-of-sample support. Our method is simple to implement, requiring only generic (and easily available) software for neural networks. We show how our approach yields good trade-offs between quality (on the one side) and speed, ease of use, genericity, generalizability, and stability (on the other side).

Many future work directions are next possible. First, we consider generalizing our approach to compute stable projections of dynamic (time-dependent) high-dimensional data and also mixed quantitative-and-qualitative data. Secondly, we consider using different network architectures, cost functions, and training procedures for more accurate handling of more complex data universes. Last but not least, we consider more refined approaches to tackle the transfer learning problem for generalizing learning from a given number of jointly considered data universes and projection techniques.


\bibliographystyle{eg-alpha-doi}
\bibliography{refs}

\end{document}